\def\year{2022}\relax
\documentclass[letterpaper]{article} 
\usepackage{aaai22}  
\usepackage{times}  
\usepackage{helvet}  
\usepackage{courier}  
\usepackage[hyphens]{url}  
\usepackage{graphicx} 
\usepackage{natbib}  
\usepackage{caption} 
\DeclareCaptionStyle{ruled}{labelfont=normalfont,labelsep=colon,strut=off} 
\usepackage{algorithm}
\usepackage{algorithmic}
\usepackage{mathtools}
\usepackage{amsmath,amssymb}
\usepackage{multirow}
\usepackage{pifont}
\usepackage{booktabs}

\usepackage{newfloat}
\usepackage{listings}
\floatstyle{ruled}
\newfloat{listing}{tb}{lst}{}
\floatname{listing}{Listing}
\nocopyright

\title{Transformer-based Network for RGB-D Saliency Detection}
\author{
    Yue Wang\textsuperscript{\rm 1},
    Xu Jia\textsuperscript{\rm 1},
    Lu Zhang\textsuperscript{\rm 1},
    Yuke Li\textsuperscript{\rm 2},
    James Elder\textsuperscript{\rm 3},
    Huchuan Lu\textsuperscript{\rm 1}
}
\affiliations{
    \textsuperscript{\rm 1}Dalian University of Technology\\
    \textsuperscript{\rm 2}UC Berkeley\\
    \textsuperscript{\rm 3}York University\\

}

\begin{document}

\maketitle

\begin{abstract}
RGB-D saliency detection integrates information from both RGB images and depth maps to improve prediction of salient regions under challenging conditions.
The key to RGB-D saliency detection is to fully mine and fuse information at multiple scales across the two modalities. 
Previous approaches tend to apply the multi-scale and multi-modal fusion separately via local operations,
which fails to capture long-range dependencies.
Here we propose a transformer-based network to address this issue.
Our proposed architecture is composed of two modules:
a transformer-based within-modality feature enhancement module (TWFEM) and a transformer-based feature fusion module (TFFM).
TFFM 
conducts a sufficient feature fusion by integrating features from multiple scales and two modalities over all positions simultaneously.
TWFEM enhances feature on each scale by selecting and integrating complementary information from other scales within the same modality before TFFM.
We show that transformer is a uniform operation 
which presents great efficacy in both feature fusion and feature enhancement, and simplifies the model design.
Extensive experimental results on six benchmark datasets demonstrate that our proposed network
performs favorably against state-of-the-art RGB-D saliency detection methods.

\end{abstract}

\section{Introduction}

\begin{figure*}[t] 
\centering
\setlength{\abovecaptionskip}{3.0pt}
\includegraphics[width=0.86\textwidth]{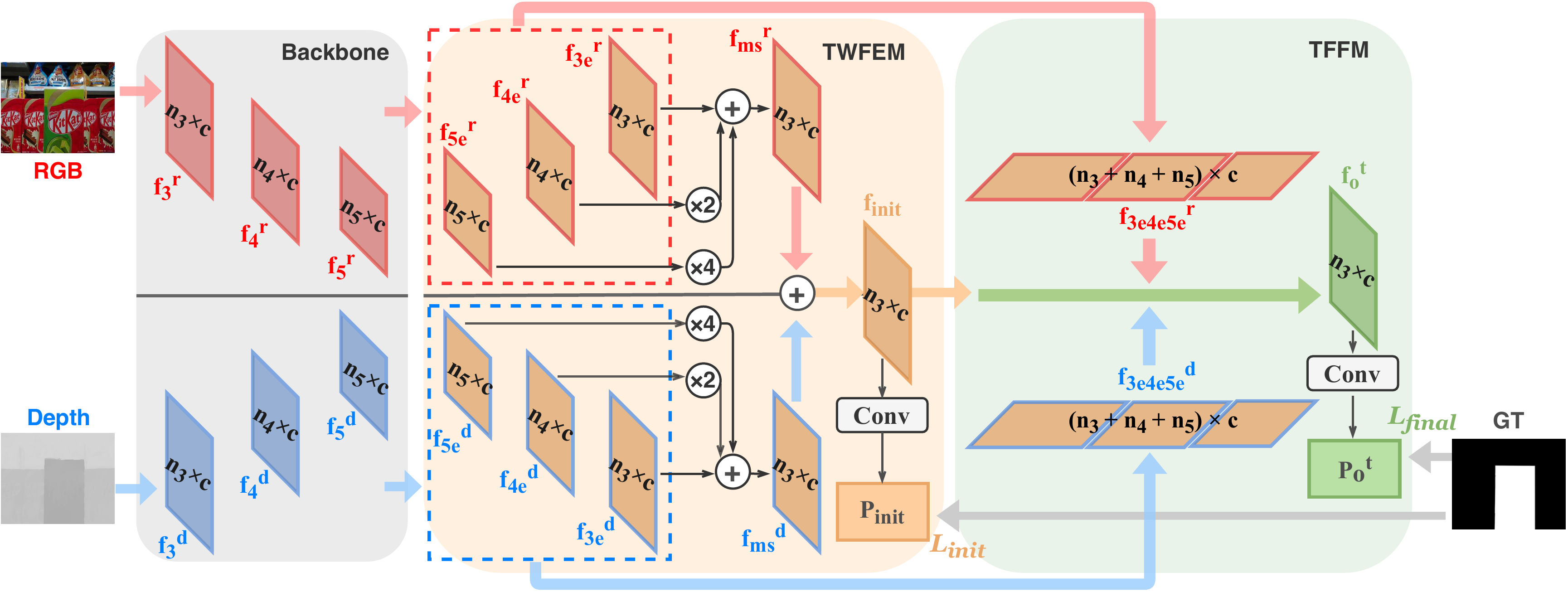}
 \caption{The structure of our overall method.
 Here, the $\times 2, \times 4$ in figure represent the $UP_{\times 2}(\cdot), UP_{\times 4}(\cdot)$ in TWFEM. }
\label{fig: flow}
  \vspace{-0.5cm} 
\end{figure*} 
 
RGB-D saliency detection aims at discovering and segmenting the most salient objects accurately in complex scenes
by exploiting both RGB image and depth data.
It can serve as an early vision step for 
visual tracking~\cite{lee2018salient}, 
object detection~\cite{xu2015show}, 
content-aware image editing~\cite{wang2018deep}, 
and image retrieval~\cite{he2012mobile}. 

RGB images capture both color cues and texture details of objects that 
may be sufficient to distinguish foreground objects from background in some simple scenes.
However, it may not work 
for highly cluttered scenes or scenes in which foreground and background share similar colors and textures.
Depth maps, on the other hand, would benefit the discovering and segmenting process 
for these complex scenes.
It records the distance of different objects in a scene to the camera, 
which provides an additional cue to capture the spatial structure and 3D layout. 
Therefore, information provided by RGB images and depth maps are complementary to each other.
Effective fusion of information from such two modalities can be critical for accurate saliency detection.

Inspired by the success of deep learning methods, 
RGB-D saliency detection apply CNN models~\cite{simonyan2014very, he2016deep, huang2017densely}
to effectively extract multi-scale features from RGB image and depth map.
High-level features from deep layers 
represent coarse-scale and semantic information in an image,
while low-level features from shallow layers
capture fine-scale details for precisely localizing boundaries of objects. 
Due to this complementarity, 
effectively fusion information across scales is also a key to successful RGB-D saliency detection.
In addition, CNN-based structures also contribute to the fusion of multi-scale multi-modal features. 
However, as a local operation, the receptive field of CNNs is limited by its kernel size and number of layers.
It makes long-range dependency modeling difficult for feature fusion in saliency detection.

There have been several methods~\cite{li2020icnet, ling2020cross, piao2019depth, chen2019three, chen2018progressively}
working on the fusion of multi-modal features across multiple scales with improved RGB-D saliency detection performance.
However, 
most of them conduct 
feature fusion
in a progressive manner by CNN-based local operations, that is, 
for each time, features from two modalities only fuse with each other within the same scale and at a local spatial neighborhood.
This could cause insufficient integration of two modalities because the potential 
long-range dependencies would be restricted.
In addition, these prior approaches tend to employ different techniques for different modules,
\textit{e}.\textit{g},
channel-wise concatenation or spatial attention for fusion across modalities; 
U-Net-like progressive integration for fusion across scales
and recurrent attention for feature enhancement.
Such diversity within modules leads to a potentially unwieldy complexity in the design of model architecture.
%
%
This motivates us to design a uniform operation applied to multi-modal multi-scale fusion and even feature enhancement, which would simplify the model design and benefit performance boost of RGB-D saliency detection.

%
%
Inspired by transformer~\cite{vaswani2017attention},
a non-local model with self attention and cross attention layer to capture long-range dependencies in an image,
we propose a novel transformer-based network for RGB-D saliency detection to address the aforementioned limitations.
It contains a Transformer-based Feature Fusion Module (TFFM) to fuse multi-scale multi-modal features globally, 
and a Transformer-based Within-modality Feature Enhancement Module (TWFEM) to enhance features 
across scales within the same modality.
Both of these modules are composed of only transformer decoders. 
And instead of only considering long-range dependencies among all positions within one feature, 
our proposed structure takes advantage of transformer's non-local essence to 
simultaneously work on features among different modalities and multiple scales. 
For TFFM, in the self-attention layer of transformer, 
a guidance for global fusion is generated by refining an initial fused feature with its self-interaction between each two positions.
In the cross-attention layer,
features from different modalities and scales are taken as embeddings of a sequence.
With such guidance,  
each position is equipped with a final fused feature that 
combines multi-modal and multi-scale information at all positions in the image,
which is further fed to a classifier to predict the saliency probability map.
In addition, TWFEM is employed within each modality to generate enhanced features fed to TFFM.
It allows feature on each scale to look through features from all other scales at all positions to 
select and integrate complementary information for enhancement while remains its original resolution.
It also simply produces the initial fused feature for TFFM.
In this way, we are able to have a uniform design based on transformers for both feature enhancement and feature fusion, which simplify the model design and boost saliency detection performance.

Main contributions of this work are three-folded.

- We introduce a novel transformer-based framework for RGB-D saliency detection which 
simultaneously 
and globally integrates features across modalities and scales.

- Our RGB-D saliency detector uses only transformers as a uniform operations 
for both feature fusion and feature enhancement, 
which shows the potential of transformer in this task and simplifies the model design.

- Extensive experiments over six benchmark datasets show that the proposed transformer-based network generally performs favorably against state-of-the-art RGB-D saliency detection methods.

\section{Related Work}

\subsection{RGB-D Saliency Detection}
Early RGB-D saliency detection methods~\cite{peng2014rgbd, cheng2014depth, zhu2017multilayer, cong2016saliency, zhu2017innovative, zhu2017three} 
focuse on hand-crafted low-level features and thus struggled to handle complex scenes.
More recent deep learning approaches extract high-level representations \cite{qu2017rgbd, han2017cnns}
and obtain multi-scale features from different levels \cite{chen2019multi} to improve the performance.
A common CNN architecture for RGB-D saliency detection 
involves a two-stream network to generate multi-scale features from two modalities, 
followed by 
several separate fusion processes via local operations.
One approach \cite{li2020icnet, ling2020cross, zhang2020select} 
first fuses multi-modal features separately within each scale and then combines the fused features across all scales.
A second approach \cite{chen2018progressively, li2020rgb, pang2020hierarchical} 
progressively merges the multi-modal features from coarse to fine scales.
Generally these approaches rely on convolutional operations which allow only information 
from adjacent spatial positions to contribute to feature fusion at one time.
To compensate for this limitation, some methods employ extra feature enhancement module.
They design 
different structures for multi-scale feature fusion,
multi-modal feature fusion and feature enhancement separately,
which complicates the network and still fail to simultaneously 
consider long-term dependencies among all scales, modalities and positions.
For example, \citet{piao2019depth} use a recurrent module to enhance the fused features, while
\citet{liu2020learning} use attention to enhance features across modalities before fusion.
Exceptions to the two-stream network approach do exist.
\citet{piao2020a2dele} use only RGB image as input to the single-stream network,
and \citet{zhao2020single} use a single-stream network that combines the RGB and depth data directly from the input.
But they also use a progressive approach for multi-scale fusion with local operations. 

\subsection{Transformers}

The transformer model is first proposed by~\citet{vaswani2017attention} for the machine translation task.
It has an encoder-decoder structure involving two kinds of attention.
The transformer encoder contains a self-attention layer,
while the transformer decoder has both self-attention and cross-attention layers.
The transformer model is adopted by~\citet{carion2020end} 
to exploit non-local relationships for end-to-end object detection.
A pixel-level feature extracted by a CNN-based backbone is refined by the self-attention layer 
in the transformer encoder module. 
The transformer decoder module then generates object-level feature from the refined encoder feature over all positions in an image.
\citet{zhu2020deformable} and \citet{zhang2020feature} show that the transformer model is capable of using multi-scale pixel-level features simultaneously to achieve better performance for object detection and segmentation tasks.
The transformer model is also used to transform multi-modal features 
for image caption and video retrieval~\cite{li2019entangled, gabeur2020multi}.

In this paper, we apply the structure of transformer 
to achieve simultaneous multi-modal and multi-scale feature fusion over all spatial positions for RGB- D saliency detection
We also show that transformer offers a unifying architecture which can also be applied for within-modality feature enhancement before fusion.

\section{Method}

In this section, we introduce the overall structure of our transformer-based network for RGB-D saliency detection.
Both our transformer-based feature fusion module (TFFM) and 
transformer-based within-modality feature enhancement module (TWFEM) 
are based upon a common transformer decoder template.
We review this template first, and then go
into the specifics of the TFFM and TWFEM. 
The overall architecture is shown in Fig.\ref{fig: flow}.

\subsection{Transformer Decoder Module}

\begin{figure}[t] 
\centering
\setlength{\abovecaptionskip}{1.0pt}
\includegraphics[width=0.38\textwidth, height=0.24\textheight]{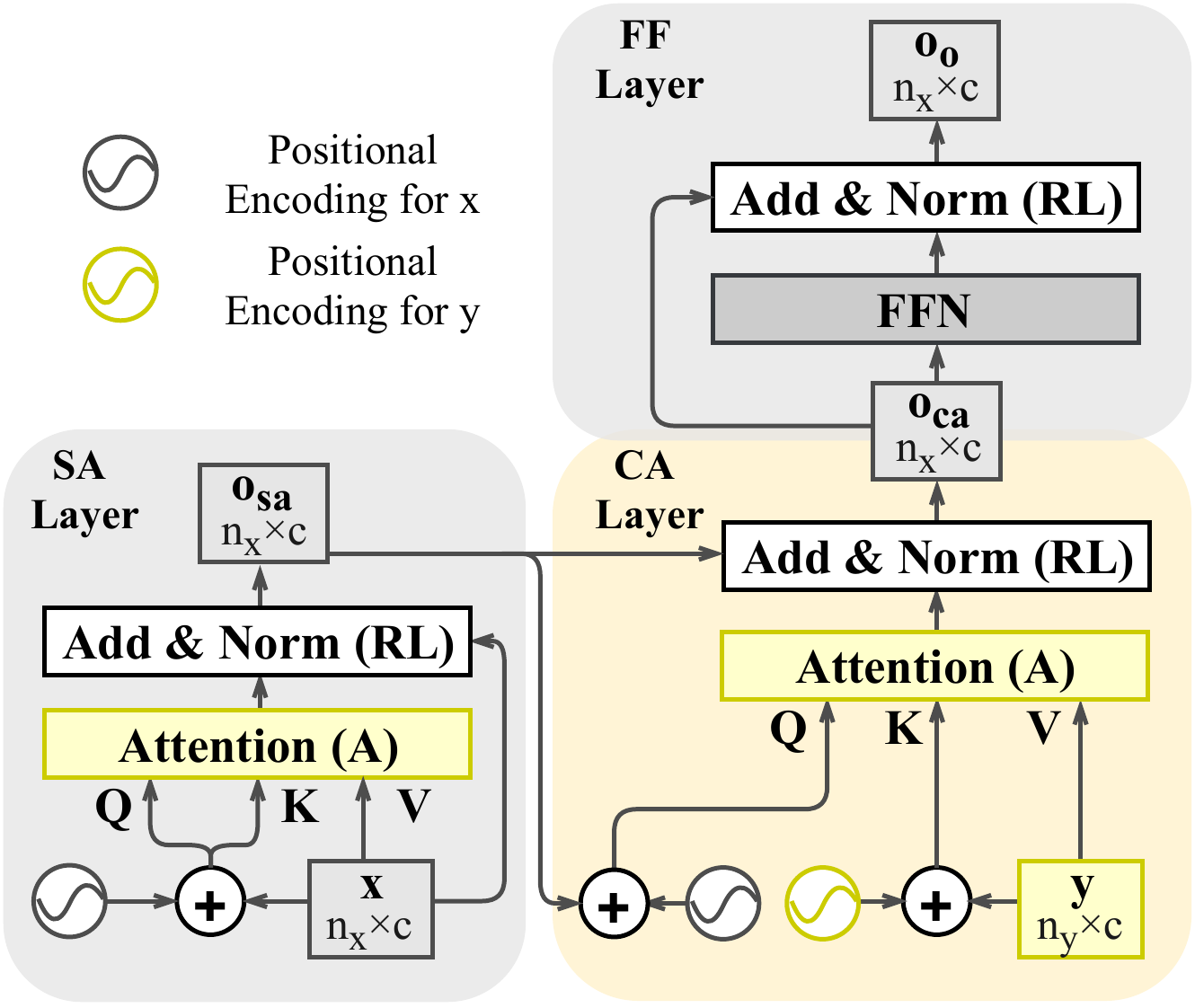}
 \caption{The structure of the transformer decoder module.
 }
\label{fig:tdm}
\vspace{-0.6cm} 
\end{figure} 

\begin{figure*}[t] 
\centering
\setlength{\abovecaptionskip}{1.0pt}
\includegraphics[width=0.93\textwidth]{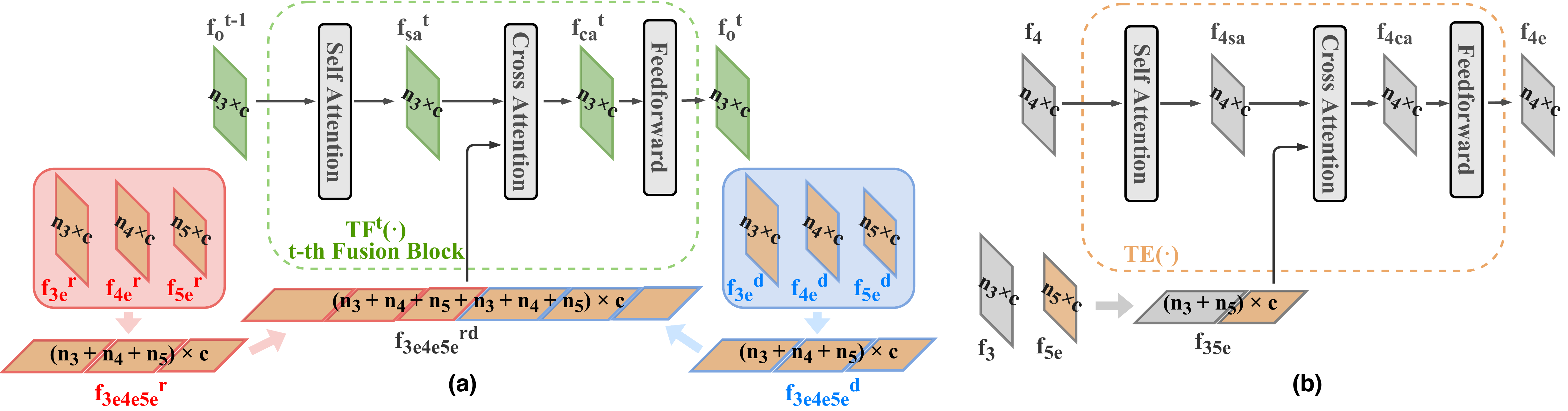}
 \caption{
 (a) The structure of our $t$-th transformer-based feature fusion block $TF^t(\cdot)$ (Eq.~\ref{tfos_s}).
 (b) The structure of our transformer-based feature enhancement block $TE(\cdot)$,
 we use the process of generating the enhanced $f_{4e}$ (Eq.~\ref{tmf4}) as example.
}
\label{fig:tfs1}
  \vspace{-0.5cm} 
\end{figure*}

The original transformer model~\cite{vaswani2017attention} consists of a transformer encoder and a transformer decoder.
Both of the TWFRM and TFFM only apply the structure of transformer decoder module.
Here, we first introduce the attention module, the core of the transformer decoder module.
Given a query $Q \in \mathbb{R}^{{n_Q}  \times {c}} $, a key $K \in \mathbb{R}^{{n_P}  \times {c}}$, and a value $V \in \mathbb{R}^{{n_P}  \times {c}}$,
the dot-product attention $A$ is computed to output a weighted sum of $V$,
with weights given by a compatibility function of $Q$ with $K$~\cite{shen2021efficient}:
\vspace{-0.0cm} 
\begin{equation}
\label{atte}
A(Q, K, V) = \rho (QK^T) V
\vspace{-0.1cm} 
\end{equation}
Here $n_Q$ is the number of positions in feature $Q$.
$K, V$ share the same number of positions $n_P$.
$c$ is the common feature dimensionality for $Q, K, V$.
$\rho(\cdot)$ represents the softmax function along each row
(see~\cite{shen2021efficient, vaswani2017attention} for details).
The output of this attention module has the same size $\mathbb{R}^{{n_Q}  \times {c}} $ as $Q$.

In this paper, we follow~\citet{vaswani2017attention} and apply the same structure of transformer decoder module.
Transformer decoder requires two input features: 
$x \in \mathbb{R}^{{n_{x}}  \times {c}}$ and $y \in \mathbb{R}^{{n_{y}}  \times {c}}$.
It is able to refine $x$ by selecting and integrating relevant information from $y$
based on the global relationship between $x$ and $y$ over all positions.
It is implemented by three sub-layers:
a self-attention (SA) layer, a cross-attention (CA) layer and a feedforward (FF) layer.
Both self-attention and cross-attention layers contain an attention module (A) and a Residual connection and Layer normalization block (RL).
Feedforward layer contains a feedforward network (FFN) with two linear transformations and a ReLU activation in between,
and a RL block.
\vspace{-0.1cm} 
\begin{equation}
\label{td}
\begin{aligned}
o_{sa} &= SA(x) = RL(A( \tilde{x},  \tilde{x}, x)) \\
o_{ca} &= CA(o_{sa}, y) = RL(A({\tilde{o}_{sa}}, \tilde{y}, y)) \\
o_{o} &= FF(o_{ca}) = RL(FFN(o_{ca}))
\end{aligned}
\vspace{-0.15cm} 
\end{equation}
$o_{sa}, o_{ca}, o_o \in \mathbb{R}^{{n_{x}}  \times {c}}$ are the outputs of self-attention, cross-attention, feedforward layers
and have the same size with input $x$.
For each attention module, $Q$ and $K$ are added by their corresponding positional encodings as~\cite{carion2020end}
and represented as $ \tilde{x}, {\tilde{o}_{sa}}, \tilde{y}$.
It help to consider the spatial information between any two positions in $Q$ and $K$ for learning the attention weight,
which is essential especially when $Q$ and $K$ have different number of positions. 
Here, the self-attention layer uses $x$ as $Q, K, V$ in attention module,
which refines $x$ based on 
its self-relationship between any two positions 
to achieve a better query $o_{sa}$ for the cross-attention layer.
The cross-attention layer uses $o_{sa}$ as $Q$ and $y$ as $K, V$ in attention module,
which looks through all positions of $y$ to extract useful information
for $o_{sa}$ to get its further refined feature $o_{ca}$. 
The final output $o_{o}$ is then calculated from $o_{ca}$ by feedforward layer.
We can name the above processes as one transformer decoder block ($TD$) and have
$o_o = TD(x, y)$.
The meanings of $x$ and $y$ are specified in TFFM and TWFEM.
And we show the overall structure of transformer decoder in Fig.\ref{fig:tdm}.

To save the memory and computational costs, we employ an efficient attention implementation~\cite{shen2021efficient} 
to replace the original attention module (A) in transformer decoder module.
Instead of multiplying $Q$ and $K$ first to form 
an ${n_Q}  \times {n_P}$ matrix,
Efficient attention ($EA$) 
first multiples $K^T$ and $V$ to form a ${c}  \times {c}$ matrix:
\begin{equation}
\label{eatte}
EA(Q, K, V) = \rho_q (Q) (\rho_k(K)^T V )
\vspace{-0.1cm} 
\end{equation}
where $\rho_q(\cdot)$ and $\rho_k(\cdot)$ are the softmax function along each row of $Q$ and each column of $K$ respectively.
Since $c$ is normally much smaller than the number of positions $n = h \times w$,
$h, w$ are the height and width for feature,
it efficiently decreases the computational costs
and still aggregates value features from all positions in a global way.
Besides, the efficient attention we employ also has the function of multi-head as~\cite{vaswani2017attention}. 

\subsection{Transformer-based Feature Fusion Module}
\label{sec:TFM}

We apply a common two-stream network as existing RGB-D saliency detection methods~\cite{piao2019depth, zhang2020select}
where each stream uses VGG16~\cite{simonyan2014very} as backbone to extract three-scale features for one modality.
Therefore, how to automatically fuse multi-modal multi-scale features
becomes the key to RGB-D saliency detection.
Different from most of the previous attempts that separately process multi-modal fusion and multi-scale fusion via local operations,
we propose a transformer-based feature fusion module (TFFM).
It manages to simultaneously fuse multi-scale multi-modal features in a global way,
where the fused feature at each positions is calculated by a adaptively weighted combination 
of information from all features at all positions.
Following~\cite{carion2020end}, our TFFM also contains 
several fusion blocks which manage to refine the fused feature gradually.

We apply the structure of transformer decoder for each fusion block.
As mentioned above, the transformer decoder module requires two inputs,
$x$ and $y$, it selects useful information from $y$ to refine $x$.
For our global feature fusion, input $y$ should contains information from all scales and modalities.
We obtain and denote features of three scales from RGB and depth streams as
$ f_{ie}^r \in  \mathbb{R}^{{n_i}  \times {c}} $ and
$ f_{ie}^d \in  \mathbb{R}^{{n_i}  \times {c}} $ respectively  (see TWFEM),
where $i \in \{3, 4, 5\}$,
$ n_i = h_i \times w_i $.
Note that these multi-scale features have different $n_i, h_i, w_i$ but share the the same dimensionality $c$.
We do not need to upsample all these features to the same size to fuse features with different resolutions together.
Instead, we directly concatenate $ \{ f_{3e}, f_{4e}, f_{5e} \} ^r$ for RGB stream to get $f_{3e4e5e}^r \in \mathbb{R}^{{n_{345}}  \times {c}}$, where $n_{345} = n_3 + n_4 + n_5$,
and $ \{ f_{3e}, f_{4e}, f_{5e} \} ^d$ for depth stream to get $f_{3e4e5e}^d \in \mathbb{R}^{{n_{345}}  \times {c}}$.
This practice is more suitable for our TFFM for two reasons.
Firstly, 
in each fusion block, we need to have access to all multi-scale multi-modal features.
Upsampling these features to a higher resolution when using them repeatedly
increases the memory and computational costs,
while our processing would save these resources by remaining features to their original sizes.
Secondly, objects in a scene have large variations in scales.
By maintaining multi-scale features to their original resolutions,
our method manage to cover objects over all scales
since features in lower resolution will still help predict objects with larger sizes,
while features in higher resolution will continue detecting small objects and boundries.
After that, we concatenate two-stream features $f_{3e4e5e}^r$ and $f_{3e4e5e}^d$ together as $f_{3e4e5e}^{rd} \in \mathbb{R}^{{(n_{345} \times 2)}  \times {c}}$. 
$f_{3e4e5e}^{rd}$ is used as $y$ to provide abundant information from all scales and modalities.

Input $x$ for each fusion block is a guidance of global feature fusion
and have the same size with the output fused feature. 
\citet{carion2020end} use a stack of transformer decoder blocks to generate object-level feature from pixel-level feature,
its $x$ should have the same size with the output object-level feature.
However, since there is no object-level prior information,
$x$ for its first block has to be set as a zero tensor 
so that the object-level feature has to be generated from zero.
While for our RGB-D saliency detection,
an initial fused feature $f_{init} \in \mathbb{R}^{{n_{3}}  \times {c}}$ with a higher resolution can be easily obtained (see TWFEM)
and used as $x$ in the first fusion block to provide a better guidance for feature fusion.
With the required $f_{init}$ and $f_{3e4e5e}^{rd}$,
the overall process of the first transformer-based fusion block is as follows:
\vspace{-0.12cm} 
\begin{equation}
\label{tm_s}
\begin{aligned}
f_{sa} &= SA(f_{init}) \\
f_{ca} &= CA(f_{sa}, f_{3e4e5e}^{rd}) \\
f_{o} &= FF(f_{ca})
\end{aligned}
\vspace{-0.12cm} 
\end{equation}
With $f_{init}$ as query, we first apply a self-attention layer to refine itself for achieving a better guidance $f_{sa}$ for fusion.
And then in the cross-attention layer, 
$f_{sa}$ on each position is refined by looking through all positions in $f_{3e4e5e}^{rd}$ to generate a better fused feature $f_{ca}$.
Therefore, for each positions in $f_{ca}$,
it contains information which is adaptively selected and fused from all positions in
multi-scale multi-modal features. 
The feedforward layer is applied to get the output fused feature $f_{o} \in \mathbb{R}^{{n_{3}}  \times {c}}$
which has the same size as $f_{init}$.
Note that the positional encoding of $f_{3e4e5e}^{rd}$ is obtained by concatenating 
positional encodings of all multi-scale multi-modal features in sequence.
Even though features in $f_{3e4e5e}^{rd}$ have different resolutions, 
this transformer-based block would still consider the spatial information during fusion.

In the following fusion blocks, input $x$ is replaced by the output $f_{o}$ from a previous block.
We name this transformer-based fusion block as $TF$
and the process of $t$-th fusion block can be written as:
\vspace{-0.1cm} 
\begin{equation}
\label{tfos_s}
f_{o}^{t} = TF^t(f_{o}^{t-1}, f_{3e4e5e}^{rd})
\vspace{-0.1cm} 
\end{equation}
where $ t \in [1, \dots, T]$, $T$ denotes the number of the straightforward fusion blocks in our TFFM.
$f_{o}^{t}$ is the output of $t$-th block. $f_{o}^{0} = f_{init}$ is the input $x$ for the first block,
$TF^t(\cdot)$ represents the process in $t$-th fusion block.
 
The proposed TFFM manages to fuse multi-scale multi-modal simultaneously in a global way.
Meanwhile, by using several fusion blocks, 
our method is able to generate the final fused feature gradually 
with information from all scales, modalities and positions.
The architecture of one transformer-based feature fusion block is shown in Fig.\ref{fig:tfs1} (a).

With $f_{o}^t \in \mathbb{R}^{{n_3} \times {c} }$ output from each block,
it is first reshaped back to form $\mathbb{R}^{{c} \times {h_{3}} \times {w_{3}} }$.
Then it is used to obtain saliency prediction $P_{o}^{t}$ by one convolutional layer as a classifier.  
For better supervision on all the fusion processes in all $T$ fusion blocks, 
$f_{o}^t$ from all blocks are used to produce $T$ saliency maps.
The overall prediction loss is calculated by:
\vspace{-0.15cm} 
\begin{equation} 
\label{initloss}
\mathcal{L}_{final} = \sum_{t=1}^{T}\mathcal{L}_{bce}(P_{o}^{t}, S)
\vspace{-0.15cm} 
\end{equation}
where $\mathcal{L}_{bce}$ stands for the binary cross-entropy loss,
$S$ is the saliency ground-truth.
And the convolutional layer is shared among all $T$ blocks for predicting the saliency maps from all $T$ fused features. 

\subsection{Transformer-based Within-modality Feature Enhancement Module}
\label{sec:TMM}

To further improve the performance,
a TWFEM is proposed to generate the enhanced multi-scale features $ \{ f_{3e}, f_{4e}, f_{5e} \}$
for each modality separately and a simple initial fused feature $f_{init}$
before TFFM.
Firstly, feature on each scale are enhanced by further extracting and integrating complementary information from other scales
within the same modality while remains its original resolution.
It is also implemented as stacks of transformer decoders
which indicates 
the flexibility of transformer decoder and uniformity of the proposed framework. 
Secondly, $f_{init}$ 
is computed by simply fusing enhanced multi-scale features of two modalities.
It is used in TFFM as a better guidance for global feature fusion.  

We first show how to generate $ \{ f_{3e}, f_{4e}, f_{5e} \}$ for each modality separately. 
For each stream, we can directly extract three-scale features from its backbone network separately
and apply one convolution layer for each feature to make them have the same dimensionality $c$,
which are denoted as 
$ f_i^r \in  \mathbb{R}^{{c}  \times {h_i} \times {w_i}} $ and 
$f_i^d \in  \mathbb{R}^{{c}  \times {h_i} \times {w_i}} $ for features from RGB stream and depth stream,
where $i \in \{3, 4, 5\}$.
This is a simple way to extract the feature on each scale for each modality,
while these features can be further enhanced by automatically selecting and integrating complementary information from other scales
since features with different scales contain different levels of information.
Features from coarser scales contain more semantic information,
while features from finer scales contain more detailed information. 
Therefore, the structure of transformer decoder is suitable for feature enhancement across scales.
It allows the enhanced features to contain richer information from different scales while remain their original resolutions.

We reshape all features to form $ f_i \in  \mathbb{R}^{{n_i}  \times {c}} $
and then enhance features within the same modality from coarse scale to fine scale in a progressive way.
It also uses the structure of transformer decoder.
For each $f_i$, we enhance it by setting itself as input $x$,
the concatenation of features from its finer scales ($f_j$
where $j>i, j \in [3,4,5]$)
and the enhanced features from its coarser scales ($f_{le}$
where $l<i, l \in [3,4,5]$)
as input $y$.
It means for each stream, 
we first enhance the coarsest-scale feature $f_5$ by $ f_3, f_4$ from the same modality while remaining $f_5$'s original resolution.
The enhancement process consists of a self-attention layer to refine $f_5$ by itself,
a cross-attention layer to extract and integrate complementary information from $ f_3 $, $ f_4$,
and a feedforward layer:
\vspace{-0.15cm} 
\begin{equation}
\begin{aligned}
f_{5sa} &= SA(f_5) \\
f_{5ca} &= CA(f_{5sa}, f_{34}) \\
f_{5e} &= FF(f_{5ca})
\end{aligned}
\vspace{-0.15cm} 
\label{tm}
\end{equation}
where $f_{34} \in \mathbb{R}^{({n_{3} + n_4)}  \times {c}}$ is formed 
by concatenating $ f_3 $ and $ f_4$.
$f_{5e} \in  \mathbb{R}^{{n_5}  \times {c}} $ is the enhanced feature of $f_5$.
Here, we also select and integrate useful information in a global way where features from all spatial positions in $ f_3 $ and $ f_4$ can be contributed to 
each position in $f_{5e}$.
We name it as a transformer-based enhancement block ($TE$)
and the above process for enhancing $f_5$ can be represented as:
\begin{equation}
\label{tmf5}
f_{5e} = TE(f_5, f_{34})
\vspace{-0.1cm} 
\end{equation}

Similarly, we can progressively compute enhanced features for the other two scales.
For $f_{4e}$, its complementary information is extracted from $f_3$ and the enhanced $f_{5e}$ for integration,
and for $f_{3e}$, it is from the enhanced $f_{4e}$ and $f_{5e}$:
\vspace{-0.1cm} 
\begin{equation}
\label{tmf4}
f_{4e} = TE(f_4, f_{35e})
\vspace{-0.1cm} 
\end{equation}
\begin{equation}
\label{tmf3}
f_{3e} = TE(f_3, f_{4e5e})
\end{equation}
\vspace{-0.05cm} 
where $f_{35e} \in \mathbb{R}^{(n_{3} + n_5) \times {c}}$ is the concatenation of $ f_{3}$ and $f_{5e}$.
$f_{4e5e} \in \mathbb{R}^{(n_{4} + n_5) \times {c}}$ is the concatenation of $ f_{4e}$ and $f_{5e}$.
The structure of our $TE(\cdot)$ 
is shown in Fig.\ref{fig:tfs1} (b).

Then, with the enhanced feature $ f_{ie} \in  \mathbb{R}^{{n_i} \times {c}} $,
$i \in \{3, 4, 5\}$ from both two modalities, 
we start to generate the simple initial fused feature $f_{init}$
by a simple way to first fuse multi-scale features by addition for each modality,
then fuse multi-modal features:
\vspace{-0.05cm} 
\begin{equation}
\label{tmfms}
\begin{aligned}
f_{ms} = f_{3e} + UP_{\times 2}(f_{4e}) &+ UP_{\times 4}(f_{5e})  \\
f_{init} = f_{ms}^r &+ f_{ms}^d
\end{aligned}
\vspace{-0.1cm} 
\end{equation}
where $UP_{\times 2}(\cdot)$, $UP_{\times 4}(\cdot)$ stand for $\times 2$, $\times 4$ upsample function,
which first reshape $f_{ie}$ to form $\mathbb{R}^{{c} \times {h_i} \times {w_i}} $,
then apply the corresponding bilinear upsampling and reshape it back to form $\mathbb{R}^{{n_3} \times {c}} $
as $f_{3e}$. 
For getting a better guidance for TFFM,
we reshape $f_{init}$ back to form $\mathbb{R}^{{c} \times {h_{3}} \times {w_{3}} }$
and apply one convolutional layer on it as a classifier to get an initial saliency prediction map $P_{init}$.
$P_{init}$ can be supervised by the saliency ground-truth $S$:
\begin{equation}
\label{initloss}
\mathcal{L}_{init} = \mathcal{L}_{bce}(P_{init}, S)
\vspace{-0.1cm} 
\end{equation}
\begin{table*}[t]
\setlength{\abovecaptionskip}{0.0pt}
\centering\caption{Results on different datasets.
We highlight the best three results in each column in 
\underline{\textbf{bold}}, \textbf{bold}
and \textit{\underline{underline}}.}
\centering
\resizebox{0.87\textwidth}{!}{%
\begin{tabular}{cc|cccccccccc|c}
\toprule
    Dataset & Metric 
& CPFP  & DMRA  & ICNet & DANet 
& CMWNet & S2MA  & A2dele 
& PGAR  & SSF & HDFNet    
& Ours \\
\hline

    \multirow{4}{*}{DES} 
& MAE   
& 0.038  & 0.030  & 0.027  & 0.028  
& 0.022  & \textit{\underline{0.021}} &  0.029  
& 0.026  & 0.026 & \textbf{0.020}   
& \textbf{\underline{0.018}} \\

          & $F_m$     
& 0.829  & 0.867  & 0.889  & 0.891  
& 0.900  & \textit{\underline{0.906}} &0.868  
& 0.880  & 0.883  & \textbf{0.919}  
& \textbf{\underline{0.921}} \\

          & $S_m$    
& 0.872  & 0.899  & 0.920  & 0.905  
& \textit{\underline{0.934}} & \textbf{\underline{0.941}} &  0.883  
& 0.913  & 0.903  & 0.932  
& 0.932  \\

          & $E_m$    
& 0.927  & 0.944  & 0.959  & 0.961  
& 0.967  & \textbf{0.974} & 0.919  
& 0.939  & 0.946 & \textit{\underline{0.973}}   
& \textbf{\underline{0.975}} \\

\hline

    \multirow{4}{*}{DUT-D} 
& MAE   
& 0.100  & 0.048  & 0.072  & 0.047  
& 0.056  & 0.044  & 0.042  
& \textit{\underline{0.035}} & \textbf{0.034} & 0.040   
& \textbf{\underline{0.030}} \\

          & $F_m$     
& 0.735  & 0.883  & 0.830  & 0.884  
& 0.866  & 0.885  & 0.891 
& \textbf{0.914} & \textbf{0.914}  & \textit{\underline{0.892}}  
& \textbf{\underline{0.923}} \\

          & $S_m$    
& 0.749  & 0.887  & 0.852  & 0.889  
& 0.887  & 0.902  & 0.884  
& \textbf{0.919} & \textit{\underline{0.914}} & 0.905  
& \textbf{\underline{0.924}} \\

          & $E_m$    
& 0.815  & 0.930  & 0.901  & 0.929  
& 0.922  & 0.935  & 0.929  
& \textit{\underline{0.950}} & \textbf{0.951} & 0.938  
& \textbf{\underline{0.954}} \\

\hline

    \multirow{4}{*}{NJUD}  
& MAE   
& 0.053  & 0.051  & 0.052  & 0.046  
& 0.046  & 0.053  & 0.051  
& \textit{\underline{0.042}}  & 0.043 & \textbf{0.037}  
& \textbf{\underline{0.036}} \\

          & $F_m$     
& 0.837  & 0.872  & 0.868  & 0.877  
& 0.880  & 0.865  & 0.873  
& \textbf{0.893} & \textit{\underline{0.885}} & \textbf{\underline{0.894}} 
& \textbf{\underline{0.894}} \\

          & $S_m$    
& 0.878  & 0.885  & 0.894  & 0.897  
& 0.903  & 0.894  & 0.868  
& \textit{\underline{0.909}}  & 0.898  & \textit{\underline{0.911}} 
& \textbf{\underline{0.913}} \\

          & $E_m$    
& 0.900  & 0.920  & 0.913  & 0.926  
& 0.923  & 0.916  &  0.916  
& \textbf{0.935} & \textit{\underline{0.934}} & \textit{\underline{0.934}} 
& 0.932  \\

\hline

    \multirow{4}{*}{NLPR}  
& MAE   
& 0.038  & 0.031  & 0.028  & 0.031  
& 0.029  & 0.030  &  0.028  
& \textbf{\underline{0.024}} & \textbf{0.026} & \textit{\underline{0.027}}   
& \textbf{\underline{0.024}} \\

          & $F_m$    
 & 0.818  & 0.855  & 0.870  & 0.865  
 & 0.859  & 0.853  & \textit{\underline{0.878}}  
 & \textbf{0.885} & 0.875 & \textit{\underline{0.878}}   
 & \textbf{\underline{0.895}} \\
 
          & $S_m$    
 & 0.884  & 0.898  & 0.922  & 0.908  
 & \textit{\underline{0.917}}  & 0.915  & 0.895  
 & \textbf{\underline{0.930}} & 0.913 & 0.916   
 & \textit{\underline{0.924}} \\
 
          & $E_m$    
 & 0.920  & 0.942  & 0.945  & 0.945 
  & 0.940  & 0.942  & 0.945  
  & \textbf{0.955} & 0.951 & \textit{\underline{0.948}}   
  & \textbf{\underline{0.960}} \\
  
         \hline
  
    \multirow{4}{*}{SIP}   
  & MAE   
  & 0.064  & 0.088  & 0.069  & \textit{\underline{0.054}}  
  & 0.062  & 0.057  & 0.070  
  & 0.055  & 0.057 & \textbf{0.050}   
  & \textbf{\underline{0.049}} \\
  
          & $F_m$     
  & 0.819  & 0.815  & 0.836  & \textbf{0.864} 
  & 0.851  & 0.854  & 0.829  
  & 0.854  & 0.850 & \textit{\underline{0.863}}  
  & \textbf{\underline{0.866}} \\
  
          & $S_m$    
  & 0.850  & 0.800  & 0.854  & \textbf{0.878} 
  & 0.867  & 0.872  & 0.826  
  & \textit{\underline{0.876}} & 0.867 & \textbf{0.878}  
  & \textbf{\underline{0.885}} \\
  
          & $E_m$    
  & 0.899  & 0.858  & 0.900  & \textit{\underline{0.917}}  
  & 0.909  & 0.913  & 0.889  
  & 0.912  & 0.913 & \textbf{0.920}  
  & \textbf{\underline{0.921}} \\
          
  \hline
  
    \multirow{4}{*}{STERE} 
  & MAE   
  & 0.051  & 0.050  & 0.045  & 0.047  
  & 0.043  & 0.051  & 0.044  
  & \textbf{0.041} & 0.044 & \textbf{\underline{0.039}}    
  & \textit{\underline{0.042}} \\
  
          & $F_m$     
  & 0.830  & 0.869  & 0.865  & 0.858  
  & 0.869  & 0.855  & \textit{\underline{0.877}} 
  & \textbf{\underline{0.880}} & 0.862 & \textbf{0.879}   
  & \textbf{\underline{0.880}} \\
  
          & $S_m$    
  & 0.879  & 0.874  & 0.902  & 0.892  
  & \textit{\underline{0.905}}  & 0.890  & 0.876  
  & \textbf{\underline{0.907}} & 0.889 & \textbf{0.906}   
  & 0.901  \\
  
          & $E_m$    
  & 0.907  & 0.926  & 0.926  & 0.926  
  & \textit{\underline{0.930}}  & 0.926  & 0.928  
  & \textbf{\underline{0.937}} & 0.927  & \textbf{\underline{0.937}}  
  & \textbf{0.934} \\
  
\hline
  
    \multicolumn{2}{c|}{Model Size (MB)} 
  & 291.9  & 238.8  & 312.2  & 106.8
  & 342.9  & 346.8  & 60.1  
  & 64.9  & 131.8  & 176.9   
  & 129.9  \\
  
	\bottomrule
    \end{tabular}%
	}
	\label{tab:exp_tab1}
  \vspace{-0.15cm} 
\end{table*}

\begin{figure*}[t] 
\centering
\includegraphics[width=0.90\textwidth]{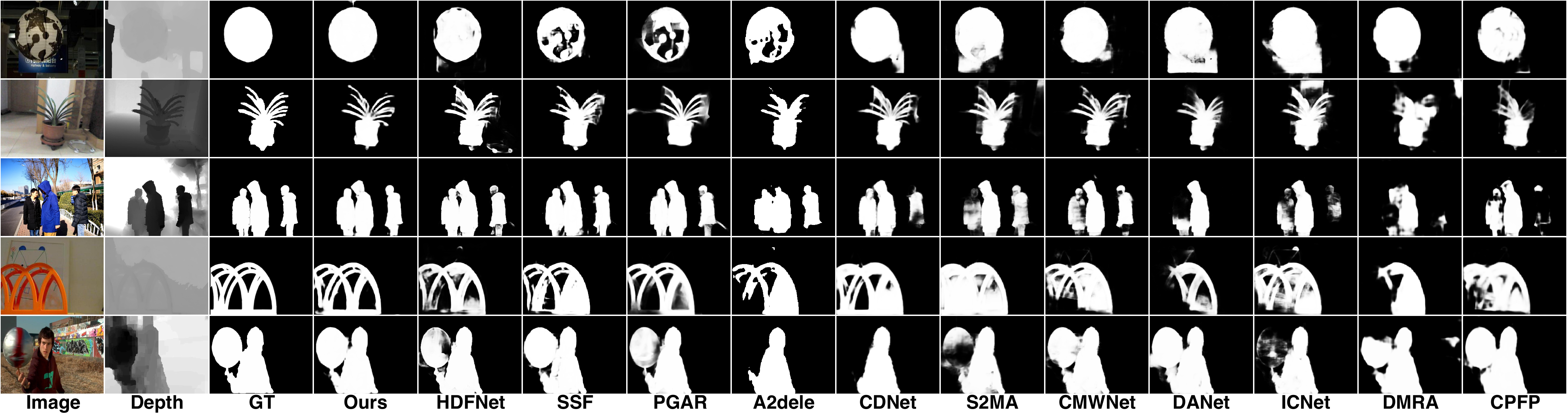}
\setlength{\abovecaptionskip}{1pt}
 \caption{Visual comparison between our method and the state-of-the-art methods.}
\label{fig:exp1}
  \vspace{-0.55cm} 
\end{figure*} 

\section{Experiments}

\begin{table*}[htbp]
\setlength{\abovecaptionskip}{-0.00pt}
  \centering \caption{Ablation study on our proposed transformer-based network. We highlight the best result in each column in 
    \underline{\textbf{bold}}.}
    \centering 
\resizebox{0.8\textwidth}{!}{%
    \begin{tabular}{c|ccc|cccc|cccc|cccc}
\toprule
         \multirow{2}{*}{Index} & \multicolumn{3}{|c}{model} & \multicolumn{4}{|c}{STERE}     & \multicolumn{4}{|c}{NLPR}      & \multicolumn{4}{|c}{DUT-D} \\
          
    &MSMMF & TWFEM & TFFM
    
    & \multicolumn{1}{c}{MAE} & \multicolumn{1}{c}{$F_m$} & \multicolumn{1}{c}{$S_m$} & \multicolumn{1}{c}{$E_m$} 
    & \multicolumn{1}{|c}{MAE} & \multicolumn{1}{c}{$F_m$} & \multicolumn{1}{c}{$S_m$} & \multicolumn{1}{c}{$E_m$} 
    & \multicolumn{1}{|c}{MAE} & \multicolumn{1}{c}{$F_m$} & \multicolumn{1}{c}{$S_m$} & \multicolumn{1}{c}{$E_m$} \\
    
    \hline
    1 &$\checkmark$ & &
    & 0.054  & 0.846  & 0.865  & 0.915  
    & 0.031  & 0.863  & 0.901  & 0.945  
    & 0.045  & 0.889  & 0.891  & 0.932  \\
    
    2 &$\checkmark$ & $\checkmark$ &
    & 0.045  & 0.861  & 0.888  & 0.925  
    & 0.028  & 0.877  & 0.913  & 0.950  
    & 0.037  & 0.901  & 0.908  & 0.942  \\
    
    3 &$\checkmark$ &  & $\checkmark$
    & 0.043  & 0.876  & 0.894  & 0.931  
    & 0.027  & 0.888  & 0.916  & 0.953  
    & 0.035  & 0.912  & 0.914  & 0.949  \\
    
    4 &$\checkmark$ & w/o prog & $\checkmark$
    & \underline{\textbf{0.042}}  & 0.873  & \underline{\textbf{0.901}}  & 0.930  
    & 0.025  & 0.887  & 0.922  & 0.955  
    & 0.032  & 0.913  & 0.921  & 0.948  \\
    
    5 &$\checkmark$ & $\checkmark$ & $\checkmark$
    & \underline{\textbf{0.042}}  & \underline{\textbf{0.880}}  & \underline{\textbf{0.901}}  & \underline{\textbf{0.934}}  
    & \underline{\textbf{0.024}}  & \underline{\textbf{0.895}}  & \underline{\textbf{0.924}}  & \underline{\textbf{0.960}}  
    & \underline{\textbf{0.030}}  & \underline{\textbf{0.923}}  & \underline{\textbf{0.924}}  & \underline{\textbf{0.954}}  \\
    
 \bottomrule     
    \end{tabular}%
    }
  \label{tab:abtab1}%
      \vspace{-0.2cm} 
\end{table*}%

\begin{table*}[htbp]
\setlength{\abovecaptionskip}{1.00pt}
  \centering
\caption{
    Comparison of setting different T for the entangled architecture. We highlight the best result in each column in \underline{\textbf{bold}}.
  }
  \centering 
\resizebox{0.70\textwidth}{!}{%
    \begin{tabular}{c|cccc|cccc|cccc}
\toprule
    
& \multicolumn{4}{|c}{STERE}  
& \multicolumn{4}{|c}{NLPR}      
& \multicolumn{4}{|c}{DUT-D} \\

    model 
    & MAE   & $F_m$  & $S_m$    & $E_m$    
    & MAE   & $F_m$  & $S_m$    & $E_m$    
    & MAE   & $F_m$  & $S_m$    & $E_m$ \\
    \hline
    
    T = 0 
    & 0.045  & 0.861  & 0.888  & 0.925  
    & 0.028  & 0.877  & 0.913  & 0.950  
    & 0.037  & 0.901  & 0.908  & 0.942  \\
    
    T = 2
    & 0.041  & 0.876  & 0.900  & 0.933  
    & 0.025  & 0.884  & 0.920  & 0.952  
    & 0.031  & 0.917  & 0.920  & 0.951  \\
    
    T = 4
    & 0.042  & \underline{\textbf{0.880}}  & 0.901  & \underline{\textbf{0.934}}  
    & \underline{\textbf{0.024}}  & \underline{\textbf{0.895}}  & \underline{\textbf{0.924}}  & \underline{\textbf{0.960}}  
    & 0.030  & \underline{\textbf{0.923}}  & 0.924  & 0.954  \\
    
    T = 5 
    & \underline{\textbf{0.040}}  & \underline{\textbf{0.880}}  & \underline{\textbf{0.905}}  & \underline{\textbf{0.934}}  
    & 0.025  & 0.891  & 0.921  & 0.956  
    & \underline{\textbf{0.029}}  & \underline{\textbf{0.923}}  & \underline{\textbf{0.926}}  & \underline{\textbf{0.956}}  \\

    \bottomrule
    \end{tabular}%
    }
  \label{tab:abtab2}%
    \vspace{-0.45cm} 
\end{table*}%

\subsection{Datasets and Evaluation Metrics}
We evaluate our proposed method on six widely used RGB-D saliency datasets including
NJUD~\cite{ju2014depth},  
NLPR~\cite{peng2014rgbd},
STERE~\cite{niu2012leveraging},
DES~\cite{cheng2014depth},
SIP~\cite{fan2019rethinking},
and DUT-D~\cite{piao2019depth}.
Following~\citet{piao2019depth}, we use the selected 800 images from DUT-D, 1485 images from NJUD and 700 images from NLPR
for training our network.
We then evaluate our model on the remaining images in these three datasets and the other three datasets. 

We adopt four widely used evaluation metrics for quantitative evaluation
including F-measure ($F_m$)~\cite{achanta2009frequency}, mean absolute error (MAE)~\cite{borji2015salient}, 
S-measure ($S_m$)~\cite{fan2017structure} and E-measure ($E_m$)~\cite{fan2018enhanced}.
In this paper, we report the mean F-measure value with the adaptive threshold as $F_m$.
For MAE, the lower value means the method is better, 
while for all other metrics, the higher value means the method is better.

\subsection{Implementation Details}
We apply PyTorch toolbox for implementation using one GeForce RTX 2080Ti GPU with 11 GB memory.
The parameters of two-stream backbone networks are both initialized by VGG16~\cite{simonyan2014very}. 
The input RGB images and depth images are all resized to $ 256 \times 256$.
We train our method for 150k iterations by ADAM optimizer~\cite{kingma2014adam}. 
The initial learning rate, batch size, number of channel $c$ and number of blocks $T$ in TFFM are set to 1e-4, 6, 128 and 4.
During evaluation, the output prediction map of the last transformer-based fusion block is used as the final saliency prediction map.

\subsection{Comparison with state-of-the-art methods}
To verify the effectiveness of our proposed transformer-based network for RGB-D saliency detection,
we compare our performance and model size with ten state-of-art RGB-D saliency detection methods 
including: 
DANet~\cite{zhao2020single},
SSF~\cite{zhang2020select},
S2MA~\cite{liu2020learning},
ICNET~\cite{li2020icnet},
CMWNet~\cite{ling2020cross},
DMRA~\cite{piao2019depth}, 
CPFP~\cite{zhao2019contrast}, 
A2dele~\cite{piao2020a2dele}, 
PGAR~\cite{chen2020progressively},
and HDFNet~\cite{pang2020hierarchical}
in Table~\ref{tab:exp_tab1}.
For a fair comparison, we report the results and model sizes with VGG16 as backbone for DANet and HDFNet.

It demonstrates that our proposed transformer-based network generally performs favorably against all the listed state-of-the-art methods. 
It benefits from our TFFM which requires a global and adaptive response of information from all scales, all modalities and all spatial positions
for our final fused feature at each position,
and our TWFEM which enhances feature on each scale with complementary information from other scales within the same modality. 
Moreover, since we 
only use the transformer-based structure with the efficient attention for both feature fusion and enhancement instead of designing diverse 
and complicated structures,
the model size of our network is smaller than almost all the other two-steam methods except for PGAR.
It is because PGAR uses a much lighter network as backbone than VGG16 for depth stream which only contains four convolution layers.
Our model size is also larger than A2dele and DANet since they apply the one-stream network.
A2dele only has RGB image as input
and DANet combines RGB images and depth maps together as one input of one-stream network. 
We also present some qualitative examples in Fig.\ref{fig:exp1}.
It shows that our method not only precisely predicts the whole saliency objects (\textit{e}.\textit{g}. row 1),
but also accurately detects the boundaries of saliency regions (\textit{e}.\textit{g}. row 4).

\subsection{Ablation Study}

\begin{figure}[b] 
  \vspace{-0.5cm} 
\centering
\includegraphics[width=0.48\textwidth]{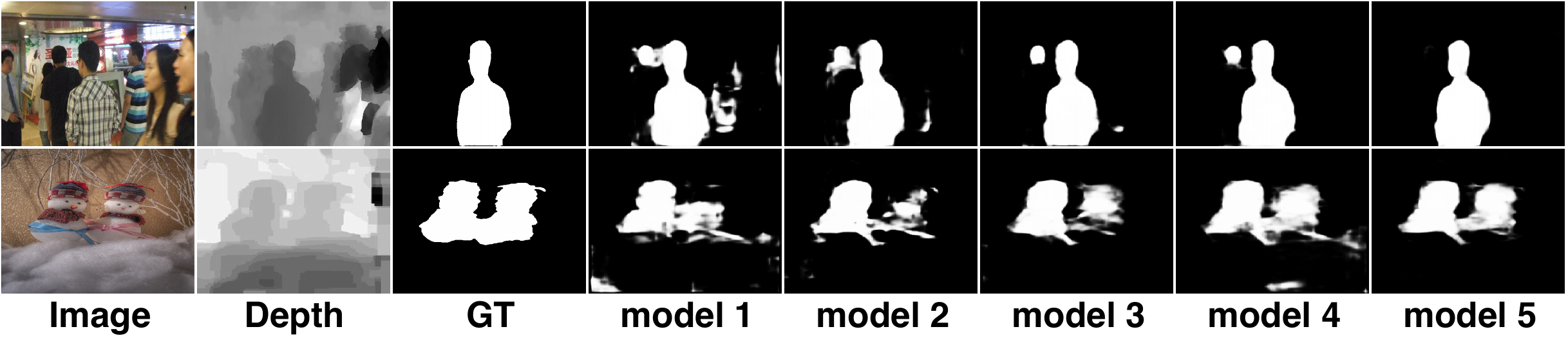}
\setlength{\abovecaptionskip}{-5pt}
 \caption{Visual examples for ablation study.}
 \setlength{\belowcaptionskip}{-5pt}
\label{fig:ab1}
  \vspace{-0.3cm} 
\end{figure} 

\textbf{Effectiveness of each transformer-based module.} 
To demonstrate the impact of each part in our overall network,
we conduct ablation studies by evaluating five subset-models and present results on three RGB-D datasets.
As shown in Table~\ref{tab:abtab1},
model $1$ which only uses MSMMF means a simple multi-scale multi-modal fusion method using~Eq.\ref{tmfms} 
in which $f_{ie}$ is replaced by $f_i$ since we do not have the enhanced features.
And the following models add TWFEM and TFFM separately to see the impact of each transformer-based module.
Model $5$ is our overall performance by using both TWFEM and TFFM.
For model $4$, we also use both TWFEM and TFFM, but in its TWFEM, 
we do not use the progressive way to get enhanced features $f_{4e}$ and $f_{3e}$ as in Eq.\ref{tmf4}, Eq.\ref{tmf3}.
Instead, we directly use $f_3, f_5$ to get $f_{4e}$ and $f_4, f_5$ to get $f_{3e}$.
We show the impact of our progressive enhanced way by the comparison between model $4$ and model $5$. 
We also include some visual examples in Fig.\ref{fig:ab1}.

It indicates that for model $1$ which only fuses multi-scale and multi-modal features in a simple way 
for saliency prediction may perform well on some simple situations.
However, without a global and adaptive fusion, it is not suitable for complex scenes
which causes poor performance on datasets like STERE.
Our TFFM can significantly improve the performance since it allows this global and adaptive fusion,
which also improves the ability of generalization for complex situations.
Our TWFEM enhances features with complimentary information across scales
which not only gains improvements for model 2 with only a simple fusion,
but also provides features with more abundant information for TFFM.
We also show that the progressively enhanced feature from coarse to fine scale in our TWFEM
can provide further improvements for saliency prediction. 

\textbf{Effectiveness of $\bold{T}$.} We also discuss the chosen number of blocks in our TFFM by conducting the experiments to set $T=0, 2, 4, 5$ for the fusion blocks in our overall transformer-based network
(by setting $T=0$ is equal to model 2 in Table~\ref{tab:abtab1}).
We present the comparisons on three datasets in Table~\ref{tab:abtab2}. 
Since the initial fused feature from our TWFEM provides useful information related to saliency prediction for our TFFM as the initial guidance. By setting different $T$ may not have a significant influence on the final result predicted by the output feature from the last block, especially when $T$ is large enough.
However, using a small number of blocks may still affect the performance on some datasets such as the NLPR dataset.
Considering both the performance and the efficiency of our model, we choose $T=4$ for our transformer-based network.

\section{Conclusion}

In this paper, we propose a transformer-based network for RGB-D saliency detection,
which uses transformer as a uniform operation for both feature enhancement and feature fusion.
It consists of two stages:
Firstly, a transformer-based single-modal feature enhancement module (TWFEM) enhances feature on each scale
by selecting and integrating complementary information from other scales within the same modality.
It also generates a simple initial fused feature.
Secondly, our transformer-based fusion module (TFFM) uses the initial fused feature as a guidance 
to simultaneously fuse multi-scale, multi-modal features.
It generates a final fused feature that at each position,
calculates by an adaptive combination of information from all scales and both modalities which captures the long-range dependencies.
Evaluation results on six RGB-D datasets demonstrate the effectiveness of our method by performing favorably against state-of-the-art RGB-D saliency methods.

\clearpage

\bibliography{egbib}

\begin{thebibliography}{45}
\providecommand{\natexlab}[1]{#1}

\bibitem[{Achanta et~al.(2009)Achanta, Hemami, Estrada, and
  Susstrunk}]{achanta2009frequency}
Achanta, R.; Hemami, S.; Estrada, F.; and Susstrunk, S. 2009.
\newblock Frequency-tuned {Salient Region Detection}.
\newblock In \emph{2009 IEEE Conference on Computer Vision and Pattern
  Recognition}, 1597--1604. IEEE.

\bibitem[{Borji et~al.(2015)Borji, Cheng, Jiang, and Li}]{borji2015salient}
Borji, A.; Cheng, M.-M.; Jiang, H.; and Li, J. 2015.
\newblock Salient {Object Detection: A Benchmark}.
\newblock \emph{IEEE Transactions on Image Processing}, 24(12): 5706--5722.

\bibitem[{Carion et~al.(2020)Carion, Massa, Synnaeve, Usunier, Kirillov, and
  Zagoruyko}]{carion2020end}
Carion, N.; Massa, F.; Synnaeve, G.; Usunier, N.; Kirillov, A.; and Zagoruyko,
  S. 2020.
\newblock End-to-end {Object Detection with Transformers}.
\newblock In \emph{European Conference on Computer Vision}, 213--229. Springer.

\bibitem[{Chen and Li(2018)}]{chen2018progressively}
Chen, H.; and Li, Y. 2018.
\newblock Progressively {Complementarity-aware Fusion Network for RGB-D Salient
  Object Detection}.
\newblock In \emph{Proceedings of the IEEE conference on computer vision and
  pattern recognition}, 3051--3060.

\bibitem[{Chen and Li(2019)}]{chen2019three}
Chen, H.; and Li, Y. 2019.
\newblock Three-stream {Attention-aware Network for RGB-D Salient Object
  Detection}.
\newblock \emph{IEEE Transactions on Image Processing}, 28(6): 2825--2835.

\bibitem[{Chen, Li, and Su(2019)}]{chen2019multi}
Chen, H.; Li, Y.; and Su, D. 2019.
\newblock Multi-modal {Fusion Network with Multi-scale Multi-path and
  Cross-modal Interactions for RGB-D Salient Object Detection}.
\newblock \emph{Pattern Recognition}, 86: 376--385.

\bibitem[{Chen and Fu(2020)}]{chen2020progressively}
Chen, S.; and Fu, Y. 2020.
\newblock Progressively {Guided Alternate Refinement Network for RGB-D Salient
  Object Detection}.
\newblock In \emph{European Conference on Computer Vision}, 520--538. Springer.

\bibitem[{Cheng et~al.(2014)Cheng, Fu, Wei, Xiao, and Cao}]{cheng2014depth}
Cheng, Y.; Fu, H.; Wei, X.; Xiao, J.; and Cao, X. 2014.
\newblock Depth {Enhanced Saliency Detection Method}.
\newblock In \emph{Proceedings of international conference on internet
  multimedia computing and service}, 23--27.

\bibitem[{Cong et~al.(2016)Cong, Lei, Zhang, Huang, Cao, and
  Hou}]{cong2016saliency}
Cong, R.; Lei, J.; Zhang, C.; Huang, Q.; Cao, X.; and Hou, C. 2016.
\newblock Saliency {Detection for Stereoscopic Images Based on Depth Confidence
  Analysis and Multiple Cues Fusion}.
\newblock \emph{IEEE Signal Processing Letters}, 23(6): 819--823.

\bibitem[{Fan et~al.(2017)Fan, Cheng, Liu, Li, and Borji}]{fan2017structure}
Fan, D.-P.; Cheng, M.-M.; Liu, Y.; Li, T.; and Borji, A. 2017.
\newblock Structure-measure: A {New Way to Evaluate Foreground Maps}.
\newblock In \emph{Proceedings of the IEEE International Conference on Computer
  Vision}, 4548--4557.

\bibitem[{Fan et~al.(2018)Fan, Gong, Cao, Ren, Cheng, and
  Borji}]{fan2018enhanced}
Fan, D.-P.; Gong, C.; Cao, Y.; Ren, B.; Cheng, M.-M.; and Borji, A. 2018.
\newblock Enhanced-alignment {Measure for Binary Foreground Map Evaluation}.
\newblock \emph{arXiv preprint arXiv:1805.10421}.

\bibitem[{Fan et~al.(2019)Fan, Lin, Zhao, Liu, Zhang, Hou, Zhu, and
  Cheng}]{fan2019rethinking}
Fan, D.-P.; Lin, Z.; Zhao, J.-X.; Liu, Y.; Zhang, Z.; Hou, Q.; Zhu, M.; and
  Cheng, M.-M. 2019.
\newblock Rethinking {RGB-D Salient Object Detection: Models, Datasets, and
  Large-scale Benchmarks}.
\newblock \emph{arXiv preprint arXiv:1907.06781}.

\bibitem[{Gabeur et~al.(2020)Gabeur, Sun, Alahari, and
  Schmid}]{gabeur2020multi}
Gabeur, V.; Sun, C.; Alahari, K.; and Schmid, C. 2020.
\newblock Multi-modal {Transformer for Video Retrieval}.
\newblock In \emph{European Conference on Computer Vision (ECCV)}, volume~5.
  Springer.

\bibitem[{Gongyang et~al.(2020)Gongyang, Zhi, Linwei, Yang, and
  Haibin}]{ling2020cross}
Gongyang, L.; Zhi, L.; Linwei, Y.; Yang, W.; and Haibin, L. 2020.
\newblock Cross-modal {Weighting Network for RGB-D Salient Object Detection}.
\newblock In \emph{European Conference on Computer Vision}. Springer.

\bibitem[{Han et~al.(2017)Han, Chen, Liu, Yan, and Li}]{han2017cnns}
Han, J.; Chen, H.; Liu, N.; Yan, C.; and Li, X. 2017.
\newblock C{NNs-based RGB-D Saliency Detection via Cross-view Transfer and
  Multiview Fusion}.
\newblock \emph{IEEE transactions on cybernetics}, 48(11): 3171--3183.

\bibitem[{He et~al.(2012)He, Feng, Liu, Cheng, Lin, Chung, and
  Chang}]{he2012mobile}
He, J.; Feng, J.; Liu, X.; Cheng, T.; Lin, T.-H.; Chung, H.; and Chang, S.-F.
  2012.
\newblock Mobile {Product Search with Bag of Hash Bits and Boundary Reranking}.
\newblock In \emph{2012 IEEE Conference on Computer Vision and Pattern
  Recognition}, 3005--3012. IEEE.

\bibitem[{He et~al.(2016)He, Zhang, Ren, and Sun}]{he2016deep}
He, K.; Zhang, X.; Ren, S.; and Sun, J. 2016.
\newblock Deep {Residual Learning for Image Recognition}.
\newblock In \emph{Proceedings of the IEEE Conference on Computer Vision and
  Pattern Recognition}, 770--778.

\bibitem[{Huang et~al.(2017)Huang, Liu, Van Der~Maaten, and
  Weinberger}]{huang2017densely}
Huang, G.; Liu, Z.; Van Der~Maaten, L.; and Weinberger, K.~Q. 2017.
\newblock Densely {Connected Convolutional Networks}.
\newblock In \emph{Proceedings of the IEEE Conference on Computer Vision and
  Pattern Recognition}, 4700--4708.

\bibitem[{Jin et~al.(2021)Jin, Xu, Han, Zhang, and Cheng}]{jin2021cdnet}
Jin, W.-D.; Xu, J.; Han, Q.; Zhang, Y.; and Cheng, M.-M. 2021.
\newblock CDNet: {Complementary Depth Network for RGB-D Salient Object
  Detection}.
\newblock \emph{IEEE Transactions on Image Processing}, 30: 3376--3390.

\bibitem[{Ju et~al.(2014)Ju, Ge, Geng, Ren, and Wu}]{ju2014depth}
Ju, R.; Ge, L.; Geng, W.; Ren, T.; and Wu, G. 2014.
\newblock Depth {Saliency based on Anisotropic Center-surround Difference}.
\newblock In \emph{2014 IEEE International Conference on Image Processing
  (ICIP)}, 1115--1119. IEEE.

\bibitem[{Kingma and Ba(2015)}]{kingma2014adam}
Kingma, D.~P.; and Ba, J. 2015.
\newblock Adam: A {Method for Stochastic Optimization}.
\newblock In \emph{ICLR}.

\bibitem[{Lee and Kim(2018)}]{lee2018salient}
Lee, H.; and Kim, D. 2018.
\newblock Salient {Region-based Online Object Tracking}.
\newblock In \emph{2018 IEEE Winter Conference on Applications of Computer
  Vision (WACV)}, 1170--1177. IEEE.

\bibitem[{Li et~al.(2020)Li, Cong, Piao, Xu, and Loy}]{li2020rgb}
Li, C.; Cong, R.; Piao, Y.; Xu, Q.; and Loy, C.~C. 2020.
\newblock RGB-D {Salient Object Detection with Cross-modality Modulation and
  Selection}.
\newblock In \emph{European Conference on Computer Vision}, 225--241. Springer.

\bibitem[{Li, Liu, and Ling(2020)}]{li2020icnet}
Li, G.; Liu, Z.; and Ling, H. 2020.
\newblock ICNet: {Information Conversion Network for RGB-D based Salient Object
  Detection}.
\newblock \emph{IEEE Transactions on Image Processing}, 29: 4873--4884.

\bibitem[{Li et~al.(2019)Li, Zhu, Liu, and Yang}]{li2019entangled}
Li, G.; Zhu, L.; Liu, P.; and Yang, Y. 2019.
\newblock Entangled {Transformer for Image Captioning}.
\newblock In \emph{Proceedings of the IEEE/CVF International Conference on
  Computer Vision}, 8928--8937.

\bibitem[{Liu, Zhang, and Han(2020)}]{liu2020learning}
Liu, N.; Zhang, N.; and Han, J. 2020.
\newblock Learning {Selective Self-mutual Attention for RGB-D Saliency
  Detection}.
\newblock In \emph{Proceedings of the IEEE/CVF Conference on Computer Vision
  and Pattern Recognition}, 13756--13765.

\bibitem[{Niu et~al.(2012)Niu, Geng, Li, and Liu}]{niu2012leveraging}
Niu, Y.; Geng, Y.; Li, X.; and Liu, F. 2012.
\newblock Leveraging {Stereopsis for Saliency Analysis}.
\newblock In \emph{2012 IEEE Conference on Computer Vision and Pattern
  Recognition}, 454--461. IEEE.

\bibitem[{Pang et~al.(2020)Pang, Zhang, Zhao, and Lu}]{pang2020hierarchical}
Pang, Y.; Zhang, L.; Zhao, X.; and Lu, H. 2020.
\newblock Hierarchical {Dynamic Filtering Network for RGB-D Salient Object
  Detection}.
\newblock In \emph{Computer Vision--ECCV 2020: 16th European Conference,
  Glasgow, UK, August 23--28, 2020, Proceedings, Part XXV 16}, 235--252.
  Springer.

\bibitem[{Peng et~al.(2014)Peng, Li, Xiong, Hu, and Ji}]{peng2014rgbd}
Peng, H.; Li, B.; Xiong, W.; Hu, W.; and Ji, R. 2014.
\newblock R{GBD Salient Object Detection: A Benchmark and Algorithms}.
\newblock In \emph{European Conference on Computer Vision}, 92--109. Springer.

\bibitem[{Piao et~al.(2019)Piao, Ji, Li, Zhang, and Lu}]{piao2019depth}
Piao, Y.; Ji, W.; Li, J.; Zhang, M.; and Lu, H. 2019.
\newblock Depth-Induced {Multi-Scale Recurrent Attention Network for Saliency
  Detection}.
\newblock In \emph{Proceedings of the IEEE International Conference on Computer
  Vision}, 7254--7263.

\bibitem[{Piao et~al.(2020)Piao, Rong, Zhang, Ren, and Lu}]{piao2020a2dele}
Piao, Y.; Rong, Z.; Zhang, M.; Ren, W.; and Lu, H. 2020.
\newblock A2dele: {Adaptive and Attentive Depth Distiller for Efficient RGB-D
  Salient Object Detection}.
\newblock In \emph{Proceedings of the IEEE/CVF Conference on Computer Vision
  and Pattern Recognition}, 9060--9069.

\bibitem[{Qu et~al.(2017)Qu, He, Zhang, Tian, Tang, and Yang}]{qu2017rgbd}
Qu, L.; He, S.; Zhang, J.; Tian, J.; Tang, Y.; and Yang, Q. 2017.
\newblock R{GBD Salient Object Detection via Deep Fusion}.
\newblock \emph{IEEE Transactions on Image Processing}, 26(5): 2274--2285.

\bibitem[{Shen et~al.(2021)Shen, Zhang, Zhao, Yi, and Li}]{shen2021efficient}
Shen, Z.; Zhang, M.; Zhao, H.; Yi, S.; and Li, H. 2021.
\newblock Efficient {Attention: Attention with Linear Complexities}.
\newblock In \emph{Proceedings of the IEEE/CVF Winter Conference on
  Applications of Computer Vision}, 3531--3539.

\bibitem[{Simonyan and Zisserman(2015)}]{simonyan2014very}
Simonyan, K.; and Zisserman, A. 2015.
\newblock Very {Deep Convolutional Networks for Large-scale Image Recognition}.
\newblock In \emph{ICLR}.

\bibitem[{Vaswani et~al.(2017)Vaswani, Shazeer, Parmar, Uszkoreit, Jones,
  Gomez, Kaiser, and Polosukhin}]{vaswani2017attention}
Vaswani, A.; Shazeer, N.; Parmar, N.; Uszkoreit, J.; Jones, L.; Gomez, A.~N.;
  Kaiser, {\L}.; and Polosukhin, I. 2017.
\newblock Attention is {All You Need}.
\newblock In \emph{Advances in Neural Information Processing Systems},
  5998--6008.

\bibitem[{Wang, Shen, and Ling(2018)}]{wang2018deep}
Wang, W.; Shen, J.; and Ling, H. 2018.
\newblock A {Deep Network Solution for Attention and Aesthetics Aware Photo
  Cropping}.
\newblock \emph{IEEE Transactions on Pattern Analysis and Machine
  Intelligence}, 41(7): 1531--1544.

\bibitem[{Xu et~al.(2015)Xu, Ba, Kiros, Cho, Courville, Salakhudinov, Zemel,
  and Bengio}]{xu2015show}
Xu, K.; Ba, J.; Kiros, R.; Cho, K.; Courville, A.; Salakhudinov, R.; Zemel, R.;
  and Bengio, Y. 2015.
\newblock Show, {Attend and Tell: Neural Image Caption Generation with Visual
  Attention}.
\newblock In \emph{International Conference on Machine Learning}, 2048--2057.

\bibitem[{Zhang et~al.(2020{\natexlab{a}})Zhang, Zhang, Tang, Wang, Hua, and
  Sun}]{zhang2020feature}
Zhang, D.; Zhang, H.; Tang, J.; Wang, M.; Hua, X.; and Sun, Q.
  2020{\natexlab{a}}.
\newblock Feature {Pyramid Transformer}.
\newblock In \emph{European Conference on Computer Vision}, 323--339. Springer.

\bibitem[{Zhang et~al.(2020{\natexlab{b}})Zhang, Ren, Piao, Rong, and
  Lu}]{zhang2020select}
Zhang, M.; Ren, W.; Piao, Y.; Rong, Z.; and Lu, H. 2020{\natexlab{b}}.
\newblock Select, {Supplement and Focus for RGB-D Saliency Detection}.
\newblock In \emph{Proceedings of the IEEE/CVF Conference on Computer Vision
  and Pattern Recognition}, 3472--3481.

\bibitem[{Zhao et~al.(2019)Zhao, Cao, Fan, Cheng, Li, and
  Zhang}]{zhao2019contrast}
Zhao, J.-X.; Cao, Y.; Fan, D.-P.; Cheng, M.-M.; Li, X.-Y.; and Zhang, L. 2019.
\newblock Contrast {Prior and Fluid Pyramid Integration for RGBD Salient Object
  Detection}.
\newblock In \emph{Proceedings of the IEEE Conference on Computer Vision and
  Pattern Recognition}, 3927--3936.

\bibitem[{Zhao et~al.(2020)Zhao, Zhang, Pang, Lu, and Zhang}]{zhao2020single}
Zhao, X.; Zhang, L.; Pang, Y.; Lu, H.; and Zhang, L. 2020.
\newblock A {Single Stream Network for Robust and Real-time RGB-D Salient
  Object Detection}.
\newblock In \emph{European Conference on Computer Vision}, 646--662. Springer.

\bibitem[{Zhu and Li(2017)}]{zhu2017three}
Zhu, C.; and Li, G. 2017.
\newblock A {Three-pathway Psychobiological Framework of Salient Object
  Detection using Stereoscopic Technology}.
\newblock In \emph{Proceedings of the IEEE International Conference on Computer
  Vision Workshops}, 3008--3014.

\bibitem[{Zhu et~al.(2017{\natexlab{a}})Zhu, Li, Guo, Wang, and
  Wang}]{zhu2017multilayer}
Zhu, C.; Li, G.; Guo, X.; Wang, W.; and Wang, R. 2017{\natexlab{a}}.
\newblock A {Multilayer Backpropagation Saliency Detection Algorithm based on
  Depth Mining}.
\newblock In \emph{International Conference on Computer Analysis of Images and
  Patterns}, 14--23. Springer.

\bibitem[{Zhu et~al.(2017{\natexlab{b}})Zhu, Li, Wang, and
  Wang}]{zhu2017innovative}
Zhu, C.; Li, G.; Wang, W.; and Wang, R. 2017{\natexlab{b}}.
\newblock An {Innovative Salient Object Detection using Center-dark Channel
  Prior}.
\newblock In \emph{Proceedings of the IEEE International Conference on Computer
  Vision Workshops}, 1509--1515.

\bibitem[{Zhu et~al.(2020)Zhu, Su, Lu, Li, Wang, and Dai}]{zhu2020deformable}
Zhu, X.; Su, W.; Lu, L.; Li, B.; Wang, X.; and Dai, J. 2020.
\newblock Deformable {DETR: Deformable Transformers for End-to-End Object
  Detection}.
\newblock \emph{arXiv preprint arXiv:2010.04159}.

\end{thebibliography}

\end{document}